%% file: main.tex
\documentclass[conference]{IEEEtran}
\IEEEoverridecommandlockouts

\input{header}

\begin{document}
\bstctlcite{IEEEexample:BSTcontrol}

\title{
    Right in Time: Reactive Reasoning in \\Regulated Traffic Spaces
}

\author{ 
    Simon Kohaut$^{1,2}$, Benedict Flade$^{3}$, Julian Eggert$^{3}$, Kristian Kersting$^{1,4,5,6}$, Devendra Singh Dhami$^{7}$
    \thanks{
        $^{1}$ Artificial Intelligence and Machine Learning Lab, \newline\hspace*{1.6em} 
         Department of Computer Science, \newline\hspace*{1.6em}
         TU Darmstadt, 64283 Darmstadt, Germany \newline\hspace*{1.6em}
        {\tt\small  firstname.surname@cs.tu-darmstadt.de}%
    }%
    \thanks{
        $^{2}$ Konrad Zuse School of Excellence in Learning and Intelligent Systems
    }%
    \thanks{
        $^{3}$ Honda Research Institute Europe GmbH, \newline\hspace*{1.6em} 
        Carl-Legien-Str. 30, 63073 Offenbach, Germany \newline\hspace*{1.6em}
        {\tt\small firstname.surname@honda-ri.de}
    }%
    \thanks{
         $^{4}$ Hessian Center for Artificial Intelligence (hessian.AI)
    }%
    \thanks{
         $^{5}$ Centre for Cognitive Science
    }%
    \thanks{
         $^{6}$ German Center for Artificial Intelligence (DFKI)
    }%
    \thanks{
        $^{7}$ Uncertainty in Artificial Intelligence Group, \newline\hspace*{1.6em}
        Department of Mathematics and Computer Science, \newline\hspace*{1.6em}
        TU Eindhoven, 5600 MB Eindhoven, Netherlands%
    }
}

\maketitle

\begin{abstract}
Exact inference in probabilistic First-Order Logic offers a promising yet computationally costly approach for regulating the behavior of autonomous agents in shared traffic spaces. 
While prior methods have combined logical and probabilistic data into decision-making frameworks, their application is often limited to pre-flight checks due to the complexity of reasoning across vast numbers of possible universes. 
In this work, we propose a reactive mission design framework that jointly considers uncertain environmental data and declarative, logical traffic regulations. 
By synthesizing Probabilistic Mission Design (ProMis) with reactive reasoning facilitated by Reactive Circuits (RC), we enable online, exact probabilistic inference over hybrid domains.
Our approach leverages the Frequency of Change inherent in heterogeneous data streams to subdivide inference formulas into memoized, isolated tasks, ensuring that only the specific components affected by new sensor data are re-evaluated. 
In experiments involving both real-world vessel data and simulated drone traffic in dense urban scenarios, we demonstrate that our approach provides orders of magnitude in speedup over ProMis without reactive paradigms. 
This allows intelligent transportation systems, such as Unmanned Aircraft Systems (UAS), to actively assert safety and legal compliance during operations rather than relying solely on preparation procedures.
\end{abstract}
\begin{IEEEkeywords}
    Mission Design, Probabilistic Logic Programs, Reactive Programming, Advanced Air Mobility
\end{IEEEkeywords}

\section{Introduction}

Integrating Unmanned Aircraft Systems (UAS) into urban environments, a concept often referred to as Advanced Air Mobility (AAM), poses a dual challenge: high-level safety compliance and low-level operational efficiency. 
To navigate human-inhabited spaces safely, autonomous agents must not only avoid physical collisions but also adhere to complex legal, spatial, and temporal regulations. 
Existing mission design frameworks, such as ProMis~\cite{kohaut2023mission, kohaut2025nesyits}, have addressed this by applying Neuro-Symbolic systems, such as variants of Problog~\cite{problog}, to formalize laws and operator constraints as symbolic, white-box models that reason over uncertain data.

ProMis generates Probabilistic Mission Landscapes (PML) that express the validity of an agent's state-space as a scalar field of probabilities, which has found application in planning~\cite{kohaut2024ceo,kohaut2025hybrid}, prediction~\cite{kohaut2025constitutional}, and control~\cite{kohaut2025coco}.
However, a significant bottleneck remains: exact inference in these complex probabilistic models often incurs prohibitive computational costs. 
This challenge is particularly acute for autonomous agents that must update their beliefs in the face of asynchronous, high-frequency streams of information. 
Not only is inference time growing exponentially with the complexity of the regulations, but sample density also becomes a concern when covering large mission areas accurately, making real-time recomputation difficult. 
While probabilistic modeling is often concerned with searching concise representations of the solution space of logical models for efficient inference, the reactive paradigm has recently received increased attention for keeping up with the dynamics of the real world.
For example, systems such as RxInfer~\cite{DBLP:journals/jossw/BagaevPV23} aim to analyse the probabilistic model itself and adapt its structure to respond quickly to changes in input data. 

\begin{figure}[t]
    \centering
    \includegraphics[width=\linewidth]{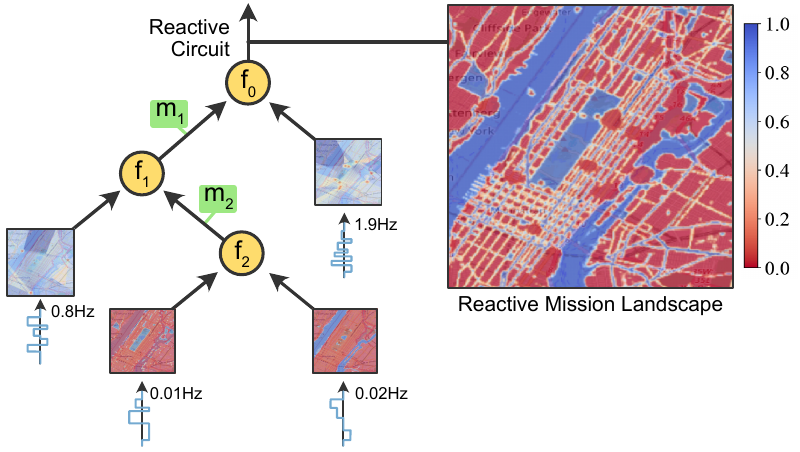}
    \caption{
        \textbf{Reasoning in Regulated Traffic Spaces via Reactive Circuits for Reactive Mission Landscapes:}
        By compiling a First-Order Logic representation into an initial Weighted Model Counting formula and then optimizing the formula's structure based on the update frequency of incoming signals, methods such as Probabilistic Mission Design can move from expensive preparation procedures to real-time applications.
    }
    \label{fig:motivation}
\end{figure}

We propose a synthesis of ProMis and asynchronous reasoning to enable Reactive Mission Landscapes (RML), an online variant of PMLs for informing safety and compliance decisions throughout a mission. 
To this end, we leverage the Resin programming language and its semantics, Reactive Circuits (RC)~\cite{kohaut2026reactive}, to accelerate ProMis' hybrid probabilistic inference architecture. 
Unlike static computation graphs, RCs are time-dynamic structures that adapt to the frequency of their inputs. 

By partitioning the mission model into individually memoized sub-formulas as illustrated in \Cref{fig:motivation}, we ensure that as few components of the inference formula as possible are invalidated and recomputed throughout an agent's lifetime.
Hence, at the core of our contributions lies an adaptive mission design framework that maintains the formal interpretability of ProMis while achieving the millisecond latency required for reactive robotics.

More specifically, we contribute:
\begin{itemize}
    \item We integrate crowdsourced environmental data, such as OpenStreetMap (OSM), with real-time sensor data, including Automatic Identification System (AIS) and Automatic Dependent Surveillance-Broadcast (ADS-B), into a transparent and adaptable mission design framework.
    \item We demonstrate Reactive Mission Landscapes (RML), an evolution of ProMis' PMLs that supports fast online belief updates over asynchronous input data to ensure safe and compliant mission design throughout an agent's lifetime, based on RC inference.
    \item We provide an open-source implementation of reactive ProMis, including interfaces for its environment representation, modeling regulated traffic spaces, and maintaining RMLs throughout the agent's mission in both global and local frames of reference.
\end{itemize}
We proceed by laying out prior work related to our methods before showing how RMLs are obtained from a synthesis of ProMis with Resin and RCs.
We demonstrate, using a mix of real-world OSM and AIS data and simulated ADS-B traffic, that RMLs can be efficiently maintained in a complex urban environment with heterogeneous traffic participants.
Finally, we discuss our findings and point towards future work.

\section{Related Work}

\subsection{Neuro-Symbolic Systems}

While early probabilistic logic languages~\cite{bayesian_logic,problog,inference_in_plp} were not formulated for end-to-end learning with artificial neural networks, languages such as DeepProbLog~\cite{deepproblog}, NeurASP~\cite{neurasp}, and SLASH~\cite{slash} close this gap and combine the strengths of neural and probabilistic reasoning. Recent work has explored the application of such neuro-symbolic approaches in robotics, notably in frameworks like Probabilistic Mission Design (ProMis)~\cite{kohaut2023mission,kohaut2024ceo}, which encode traffic laws using hybrid probabilistic first-order logic~\cite{nitti2016probabilistic}.

However, inference in these frameworks traditionally relies on static computation graphs. This poses a severe limitation for autonomous agents deployed in highly dynamic environments, as the entire logical theory must be re-evaluated whenever a single sensor reading changes. To address the dynamics of the real world, the reactive reasoning paradigm has recently gained traction. Systems like RxInfer~\cite{DBLP:journals/jossw/BagaevPV23} adapt the structure of probabilistic models to respond quickly to streaming data. Building on this momentum, the proposed architecture introduces Reactive Circuits to the spatial-legal domain to bypass the limitations of static inference.

\begin{figure*}
    \centering
    \includegraphics[width=\linewidth,trim={0cm 5cm 0cm 5cm},clip]{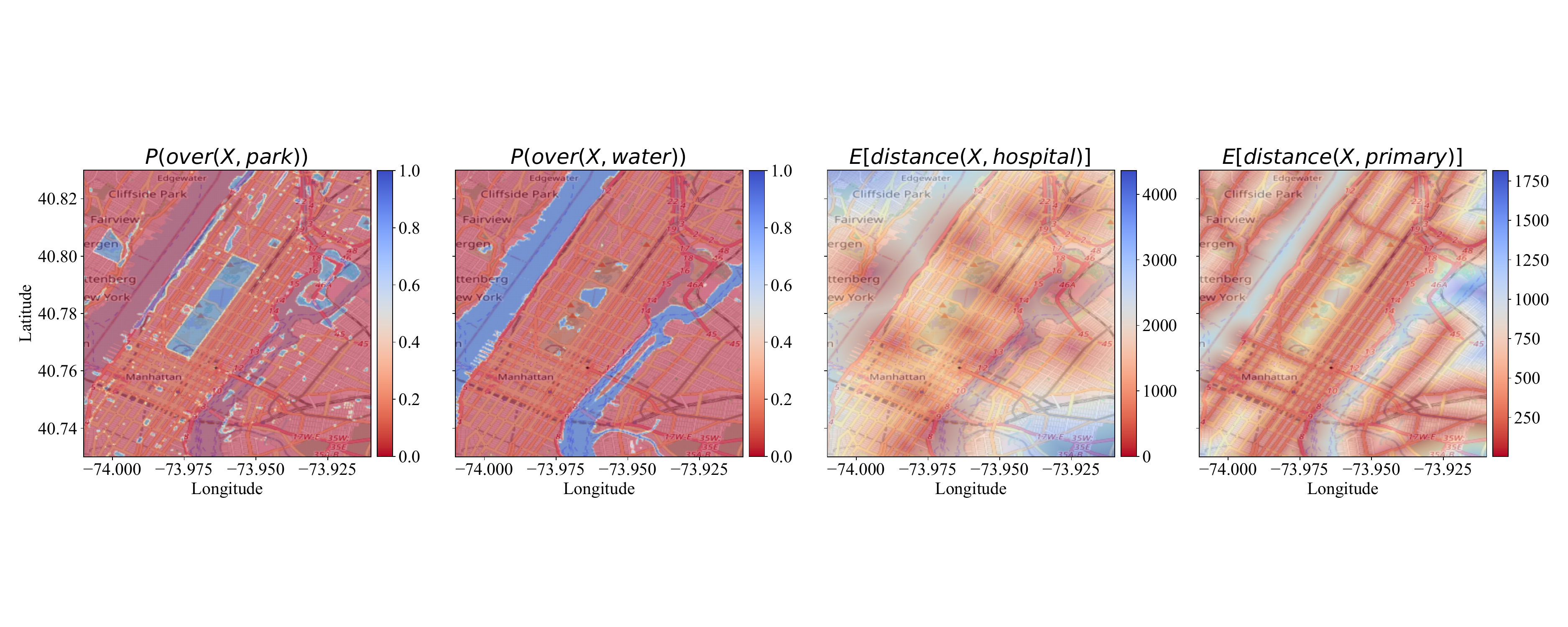}
    \caption{
        \textbf{Quasi-static statistical spatial relations in complex urban environments:}
        We precompute high-resolution statistical spatial relations from crowd-sourced map data in order to provide a precise basis for mission design reasoning.
        In contrast to relations that depend on dynamic signals (see \Cref{fig:dynamic_relations}), these parameters do not need to be updated frequently and can incorporate more sophisticated statistical evaluation and refinements.
        Hence, in an RC, they will gather at the formula nodes furthest away from the root node (e.g., $f_2$ in \Cref{fig:motivation}).
    }
    \label{fig:static_relations}
\end{figure*}

\subsection{Environment Representation}
\label{sec:related:env_repr}

Reasoning about an agent's actions requires suitable environment representations that model the mission space, obstacles, and viable paths, along with the underlying uncertainties stemming from sensor noise or neural perception. Such representations can be discrete, such as Occupancy Grid Maps~\cite{thrun2002probabilistic,jun2003probability}, which have been applied, e.g., in SLAM~\cite{grisetti2005improving}, tracking~\cite{chen2006dynamic}, and path planning~\cite{himmelsbach2008lidar}. 
Alternatively, representations may be continuous, modeling the environment as scalar fields, which can be estimated using Gaussian Process (GP) regression to capture field intensities in regions of uncertainty, as shown by Qureshi et al.~\cite{qureshi2024scalar}. 
Both types of maps allow the representation of the physical environment while modeling uncertainty in sensing or occupancy. 
However, less attention has been paid to representing and enforcing high-level legal and behavioral constraints directly within the environment.

While some logic-based representations allow for reasoning about temporal rules and agent behavior~\cite{TIGER2020325}, their uncertainty representation often focuses on the agent itself rather than the environment. 
To bridge this gap, Statistical Relational Maps (StaR Maps) have been proposed~\cite{flade2023star} as a hybrid probabilistic environment representation, employing both categorical and continuous distributions to model spatial relations in noisy maps and interface with neuro-symbolic systems. 
In this work, we employ StaR Maps both for the representation and reasoning about quasi-static environment features and for features in motion, such as air and water surface vehicles.

\subsection{Probabilistic Robotics}
In navigating uncertain environments, Bayesian filters, such as the Kalman Filter family~\cite{kalman1960new, julier1995new, julier1997new}, are often used. 
Frameworks like the Constitutional Filter have been shown to extend Probabilistic Mission Design towards learning and leveraging trust into other agents in shared traffic spaces for Bayesian estimation~\cite{kohaut2025constitutional}.

Mindful planning is crucial when defining a mission as it involves generating a collision-free path to a target location while considering various cost functions under kinodynamic constraints~\cite{yang2014literature,Probst2021}. 
Well-known methods for path planning range from graph-based approaches that search for optimal and shortest path solutions, such as A*~\cite{hart1968formal,musliman2008implementing}, to sampling-based methods such as Rapid Random Trees~\cite{yang2008real}, and evolutionary hybrid approaches~\cite{hohmann2021hybrid,hohmann2022multi}.

Recent literature has bridged planning and probabilistic logic through Probabilistic Mission Landscapes (PMLs), which serve as cost landscapes for navigation under spatial and legal uncertainty in combination with objectives on physical effectiveness~\cite{kohaut2025hybrid}. 
However, extending PMLs from pre-flight planners to active, real-time monitors remains a challenge due to the inference latency inherent in evaluating hybrid probabilistic models. 
Unlike prior works that constrain these landscapes to offline pathfinding or focus on internal state estimation, the proposed approach optimizes the structural evaluation of the landscapes themselves, transitioning static mission landscapes into highly reactive, online operational safeguards.

\section{Reactive and Probabilistic Mission Design}
\label{sec:methods}

This section disseminates the formal considerations for integrating the Resin language and Reactive Circuits (RC) with the Probabilistic Mission Design (ProMis) framework.

\subsection{Modeling Regulated Traffic Spaces in Asynchronous\\Probabilistic Logic Programs}
Hybrid Probabilistic Logic Programs (HPLP) leverage a combination of First-Order Logic (FOL) and discrete and continuous distributions to model reasoning tasks with heterogeneous sources of uncertainty. 
Resin as a reactive HPLP integrates these data sources and inferred targets as channels with an identifying token and type annotation:
\begin{align*}
    s &\leftarrow source(c, t).\\
    d &\rightarrow target(c).
\end{align*}
signifying that the ground atom $s$ is updated via channel $c$ with values of domain $t$, for which we will consider either discrete \textit{Probability} or continous \textit{Density} channels.

ProMis can leverage environment representations such as Statistical Relational Maps (StaR Maps)~\cite{flade2023star} to represent spatial relations, such as $distance(X, T)$ and $over(X, T)$, as location-dependent distributions. 
To do so, StaR Maps project earth-referenced features such as crowd-sourced OpenStreetMap data into a local navigation space to sample the respective relation from a stochastic error model~\cite{flade2021error}.
Since Resin and RCs are fully vectorized, we can omit the location variable $X$.

\subsection{Time-dynamic Spatial Relations}
To bridge the gap between static logical models and the asynchronous nature of regulated traffic environments, our architecture integrates StaR Map spatial relations systematically as Resin source signals.
Namely, we generate for each StaR Map relation code akin to
\begin{align*}
    over(park) &\leftarrow source("/over/park", Probability).\\
    distance(uas) &\leftarrow source("/distance/uas", Density).
\end{align*}

We consider quasi-static relations, e.g., derived from OpenStreetMap, which rarely change during a mission. 

For example, as illustrated in \Cref{fig:static_relations}, occupancy relations such as $over(park)$ are mapped to Probability signals where the degree of belief $P(over(park))\in[0,1]$ is derived in StaR Maps via moment matching. 
Similarly, spatial distances to fixed infrastructure, such as $distance(hospital)$, are mapped to Density signals. 
These quasi-static parameters enable high-density sampling as a preparatory step without burdening the real-time inference loop.

\begin{figure}
    \centering
    \includegraphics[width=\linewidth,trim={0cm 5cm 0cm 5cm},clip]{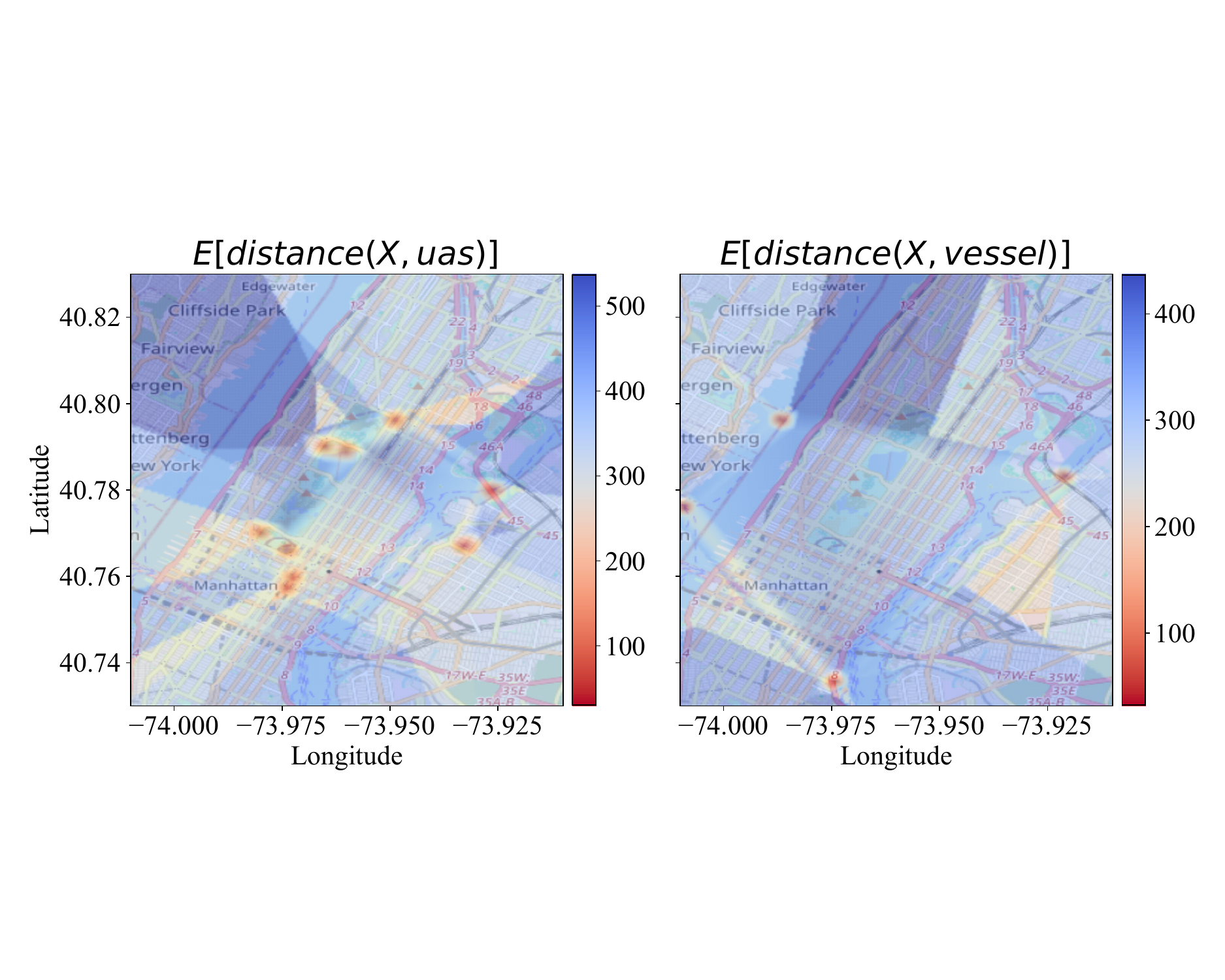}
    \caption{
        \textbf{Dynamic statistical spatial relations in multi-modal traffic environments:}
        While statistical spatial relations based on, e.g., OpenStreetMap data, are quasi-static and can be prepared at high resolution beforehand, we sample the parameters of dynamic relations, such as agent distances around the reported expected locations, and apply both linear and nearest-neighbour interpolation.
    }
    \label{fig:dynamic_relations}
\end{figure}

Furthermore, we consider real-time traffic states as dynamic spatial relations that result in frequent, but asynchronous, updates. 
For relations with continuous uncertainty tracking moving entities, such as $distance(uas)$ or $distance(vessel)$ as shown in \Cref{fig:dynamic_relations}, the output is dynamically mapped and updated as a Density signal type.
By sampling the parameters of these dynamic relations around the reported expected locations and applying both linear and nearest-neighbor interpolation, the system maintains accurate localized uncertainty models.

\subsection{Reactive Mission Landscapes}
For ProMis, we assume a single \textit{landscape} target signal to be present in the program.
Solving the Resin program for said target then induces a Weighted Model Counting formula.

Hence, the probability of the Reactive Mission Landscape (RML) is calculated as the sum of the probabilities of its models $j\in\mathcal{J}$:
\begin{equation}
    P(landscape)=\sum_{j\in\mathcal{J}}\prod_{s\in\mathcal{S}}P(s=j(s))
\end{equation}
RCs adapt this formula by partitioning it in a graphical fashion as illustrated in \Cref{fig:motivation}.
That is, RCs are Directed Acyclic Graphs over formula nodes $\mathcal{F}$, source signals $\mathcal{S}$, and memoized values $\mathcal{M}$. 
This allows the RC to autonomously adapt to the volatility of its input signals, quantified by the Frequency of Change (FoC) $\lambda_{\phi,n}(t)$. 
Updates are processed through a reactive evaluation scheme that re-computes only the invalidated ancestors $Dep(s_{n})$ of a modified signal. 
Hence, one obtains an efficiency gain
\begin{equation}
    \rho_{GAIN}(t)=\frac{\sum_{n}\lambda_{\phi,n}(t)\Omega}{\sum_{n}\lambda_{\phi,n}(t)\sum_{i\in Dep(s_{n})}\omega_{i}}
\end{equation}
where $\Omega$ represents the operations for a full evaluation and $\omega_{i}$ the number of operations for a specific formula node $f_{i}$.

\begin{figure}[t]
    \centering
    \includegraphics[width=0.7\linewidth,trim={2cmcm 0.75cm 0cm 0.75cm},clip]{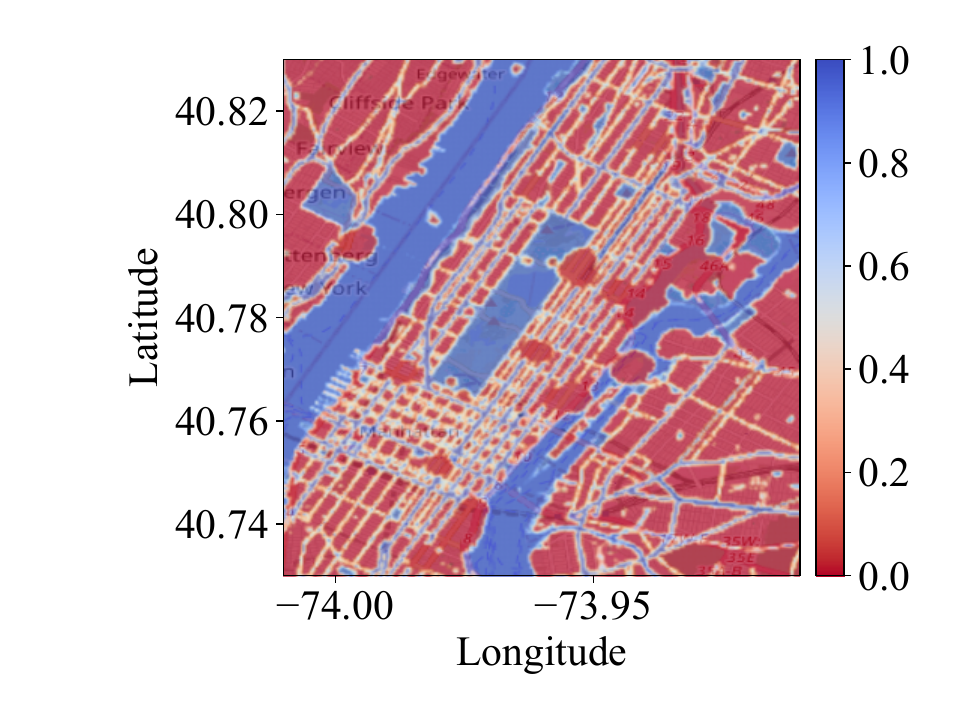}
    \caption{
        \textbf{Large-scale and real-time mission design in dense, urban cores:}
        We show the resulting mission landscape as a scalar field of probabilities of satisfying all mission requirements (listed in \Cref{listing:resin}) across an approximately $\SI{64}{\kilo\meter\squared}$ area spanning across New York City, including parts of the East and Hudson rivers.
    }
    \label{fig:landscape}
\end{figure}

The distinct temporal behaviors of the signals outlined above mandate an adaptive inference structure. 
The RC subdivides its inference formula by clustering signals based on their Frequency of Change $\lambda_{\phi,n}(t)$. 
More specifically, $k \cdot h \leq \lambda_{\phi,n}(t) < (k+1) \cdot h \implies c(s_n) = k$, where $c(s_n)$ is the assigned cluster, $h$ is the partition width, and $\phi$ is an indicator function that decides if an incoming message carried a meaningful update (e.g., based on a simple distance between the old and a new value). 
This triggers restructuring operations that preserve the semantics of the mission laws while optimizing the graph structure to maximize memoization.

The target signal $landscape$ is then computed reactively. 
That is, upon a meaningful update to a spatial signal $s_{n}$ (e.g., a drone moves, changing the pre-occupied airspaces), the system re-evaluates \textit{only} the subset of formula nodes $f_{i}\in Dep(s_{n})$:
\begin{equation}
    m_{i}(t)=f_{i}(\{s_{j}(t)|s_{j}\in C(f_{i})\}, \{m_{k}|m_{k}\in C(f_{i})\})
\end{equation}
For an in-depth discussion of the mathematical properties of RCs, we  refer to the original publication~\cite{kohaut2026reactive}.

\begin{listing}[t]
\caption{
    \textbf{Reactive and Probabilistic Mission Design}:
    This Resin program illustrates how statistical spatial relations are integrated to reason over admissible airspaces for an autonomous UAS in a dense urban environment with multiple modes of mobility.
}%
\label{listing:resin}%
    \centering
    \begin{minted}
    [
        frame=none,
        autogobble,
        fontsize=\footnotesize,
        xleftmargin=20pt,
        linenos
    ]{python}
        # Quasi-static data from OpenStreetMap
        over(park) <- source(
            "/over/park", Probability
        ).
        distance(primary) <- source(
            "/distance/primary", Density
        ).
        ...

        # Frequently updated data from AIS and ADS-B
        distance(vessel) <- source(
            "/distance/vessel", Density
        ).
        distance(uas) <- source(
            "/distance/uas", Density
        ).

        # Exemplary permitted airspaces 
        permitted if over(water).
        permitted if over(park).
        permitted if distance(primary) < 35.
        permitted if distance(primary) < 15.
    
        # Keeping safe distances to infrastructure
        building_safety if distance(hospital) > 200.

        # Keeping safe distances to mobile entities
        agent_safety if 
            distance(vessel) > 100 and
            distance(uas) > 100.
        
        # The Reactive Mission Landscape
        landscape if 
            permitted and 
            building_safety and 
            agent_safety.

        landscape -> target("/landscape").
    \end{minted}
\end{listing}

\newpage
\section{Experiments}

The evaluation focuses on demonstrating the applicability of the ProMis framework to real-time mission design and on its computational performance when coupled with Resin as encoding of regulations and RCs for inference. 
Hence, we aim to answer the following questions: 
\begin{enumerate}[\textbf{(Q1)}]
    \item[\textbf{(Q1)}] Can we represent traffic regulations in a dense urban environment with different types of static and dynamic features using a combination of StaR Maps and Resin?
    \item[\textbf{(Q2)}] Can Resin and RCs capture the different dynamics inherent to mission design tasks in urban environments? 
    \item[\textbf{(Q3)}] Can RCs maintain up-to-date RMLs within a dynamic, large-scale, urban environment, where both simulated ADS-B data from Unmanned Aircraft Systems (UAS) and real-world AIS data from vessels need to be considered?
\end{enumerate}
All experiments were conducted on an Intel Core i7-9700k desktop CPU with 32GB of memory.
We report the means and standard deviations obtained from multiple runs and seeds.

Data distribution between processes, e.g., the StaR Map environment representation and RCs for reasoning, is facilitated by employing ROS2 as a middleware~\cite{ros2}.

\subsection{Reactive Probabilistic Mission Design in New York City}
The experimental setup uses a simulation environment in New York City with map data from OpenStreetMap (OSM), AIS data from NOAA~\cite{marinecadastre_vesseltraffic}, and simulated UAS traffic across approximately $\SI{64}{\kilo\meter\squared}$ of mission space.
For the simulated part of the environment, we dedicate three vertiports across New York City to spawn UAS that first travel to a random destination within the mission area, and then move to a randomly selected vertiport before taking on a new journey.
We created a set of AAM regulations in Resin as shown in \Cref{listing:resin}.
While the StaR Map parameters obtained from OSM have been shown in \Cref{fig:static_relations,fig:dynamic_relations}, we show the resulting RML in \Cref{fig:landscape}.

\begin{figure}[t]
    \centering
    \includegraphics[width=\linewidth]{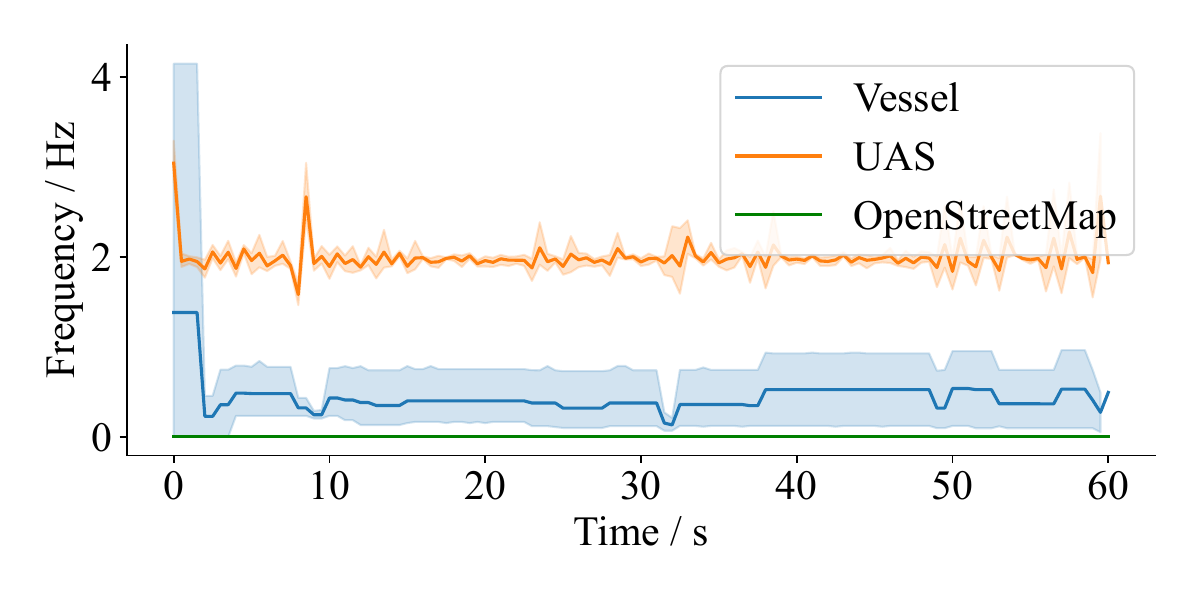}
    \caption{
        \textbf{Source signal frequencies in reactive reasoning:}
        While the map-related signals (as illustrated in \Cref{fig:static_relations}) stay constant for prolonged periods of time in the relevant area, the real-world AIS data and simulated ADS-B data arrive at different update frequencies.
        Hence, RCs can capture and leverage this dynamic and facilitate appropriate subdivisions of the reasoning formula.
    }
    \label{fig:frequencies}
\end{figure}

\subsection{Frequency Tracking and Adaptation}
The source frequencies are observed by RCs via a set of Kalman filters consuming a monotonic clock as their sensor.
We are streaming AIS data in real-time, i.e., using the timestamps they were originally sent with.
All simulated UAS move at a constant speed of $\SI{100}{\kilo\meter\per\hour}$ and report their locations at $2$~Hz.
Since the StaR Map is otherwise static, e.g., no updates to the roads or buildings of the city are performed during the experiments, no updates regarding these spatial relations are sent.
For our experiments, we define a meaningful update as a value that differs from the value in an arriving message by more than $0.003$.
The resulting mean and standard deviations of the estimated update frequencies are shown in \Cref{fig:frequencies}.

\subsection{Performance Results}

\begin{figure}[t]
    \centering
    \includegraphics[width=\linewidth]{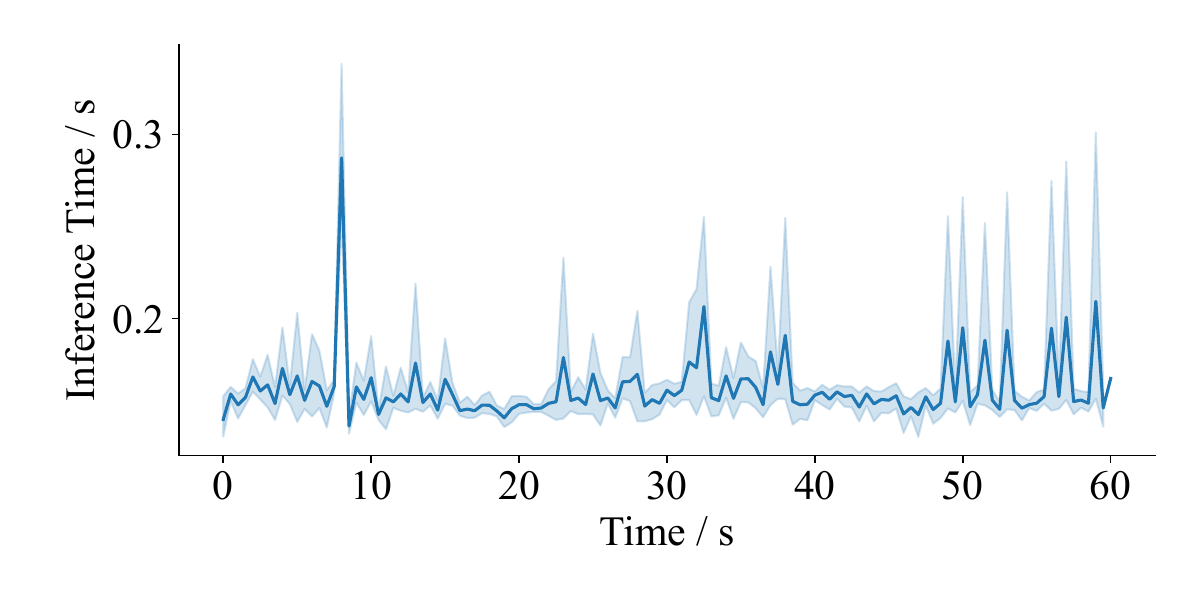}
    \caption{
        \textbf{Reactive Circuit enabled real-time inference:}
        Due to their vectorization, formula separation, and memoization based on the different emerging update frequencies (see \Cref{fig:frequencies}), RCs enable real-time updates to the Reactive Mission Landscape shown in \Cref{fig:landscape}.
        In contrast, \textbf{ProMis without RCs instead runs at approximately $\SI{42}{\second}$ per iteration}, limiting the task to pre-flight considerations. 
    }
    \label{fig:runtime}
\end{figure}

We show the mean and standard deviation of the RCs runtime for the entire $200 \times 200$ locations of the landscape in \Cref{fig:runtime}.
It can be seen that, in its configuration, adapted to the frequencies shown in \Cref{fig:frequencies}, the RC is able to provide updated RMLs at approximately $10$~Hz, with spikes in runtime occurring whenever new AIS data is obtained.
Most critically, when running ProMis without RCs as described in the original paper, each landscape update takes approximately $\SI{42}{\second}$, underscoring how RCs dramatically improve runtime and represent a big step toward real-time mission design.

\section{Conclusion}
In this work, we presented reactive ProMis, a novel synthesis of probabilistic mission design and asynchronous reasoning tailored for navigating regulated traffic spaces. 
By integrating the formal modeling of traffic regulations with the high-performance execution of RCs, we addressed the computational bottleneck traditionally associated with exact inference in complex robotic environments. 
Our architecture leverages transforming static map knowledge and dynamic sensor perception into a unified framework that optimizes computations online based on the asynchronous input frequencies.

Despite the significant speedups achieved, certain limitations remain. 
The current implementation relies on a fixed discretization of the navigation space, which may lead to inaccuracies in scenarios requiring extremely fine-grained spatial resolution.
Furthermore, while Reactive Circuits efficiently handle asynchronous updates, the initial compilation of complex Answer Set Programs into algebraic structures can still be computationally intensive for very large-scale logical theories. 
Finally, the system's reliance on crowdsourced map data introduces inherent uncertainties that, although modeled probabilistically, may still affect the reliability of the derived mission landscapes in poorly mapped regions.

Future research may explore integrating reactive ProMis with neural components such as Answer Set Networks~\cite{DBLP:journals/corr/skryagin2024asn} to scale to very large and complex logical encodings, or explore, beyond purely associative probabilistic modeling, how to model causal relations in the time-series data consumed by RCs~\cite{DBLP:journals/corr/poonia2025granger}.
While Graph Neural Network-based methods, such as Answer Set Networks, can be leveraged as solvers, supporting RC inference with such methods may further require research, e.g., for avoiding over-smoothing in classification tasks~\cite{DBLP:conf/nips/skryagin2024cna}.
One may further expand the framework's "computational plasticity" by incorporating heterogeneous computation models into the circuit's formula nodes, allowing the system to adapt not only its structure but also its internal reasoning mechanisms based on sub-formula complexity. 

\section*{Acknowledgements}
Simon Kohaut gratefully acknowledges financial support from the Honda Research Institute Europe~(HRI-EU) and the Konrad Zuse School of Excellence in Learning and Intelligent Systems (ELIZA).
This work has benefited from the Cluster of Excellence "Reasonable AI" funded by the German Research Foundation (DFG) under Germany’s Excellence Strategy—EXC-3057. 
The TU Eindhoven authors received support from their Department of Mathematics and Computer Science and the Eindhoven Artificial Intelligence Systems Institute.
Map data \copyright~OpenStreetMap contributors, licensed under the ODbL and available from \href{https://www.openstreetmap.org}{openstreetmap.org}.

\bibliographystyle{IEEEtran}
\bibliography{IEEEabrv,main.bib}

\end{document}

%% file: header.tex
\usepackage{censor}
\usepackage{microtype}
\usepackage{url}
\usepackage[hidelinks]{hyperref}
\usepackage[utf8]{inputenc}
\usepackage[small]{caption}
\usepackage{graphicx}
\usepackage{booktabs}
\usepackage{algorithm}
\usepackage{algorithmic}
\usepackage{array}
\usepackage{verbatim}
\usepackage{amsmath,amssymb,amsfonts}
\usepackage{mathtools}
\usepackage[dvipsnames]{xcolor}
\usepackage{balance}
\usepackage[normalem]{ulem}
\usepackage{comment}
\usepackage{csquotes}
\usepackage{subcaption}
\usepackage{bm}
\usepackage{siunitx}
\usepackage[formats]{listings}
\usepackage[nameinlink]{cleveref}
\usepackage{enumerate}
\usepackage{pifont}
\usepackage[frozencache,cachedir=.]{minted}
\floatstyle{plain} 
\newfloat{listing}{htbp}{lop}
\floatname{listing}{Listing}

\crefname{subfigure}{Subfigure}{Subfigures}
\Crefname{subfigure}{Subfigure}{Subfigures}